\documentclass[12pt]{article}
\usepackage{amsmath}
\usepackage{graphicx}
\usepackage[encapsulated]{CJK}
\usepackage{CJKutf8} 
\usepackage{natbib}
\usepackage{hyperref}
\usepackage{booktabs}
\usepackage{multirow}
\usepackage{float}
\usepackage{longtable}
\usepackage{geometry} 
\usepackage{caption}

\title{A Multi-LLM-Agent-Based Framework for Economic and Public Policy Analysis\footnote{We would like to thank DAI Kaiyan for technical assistance and the seminar participants at the School of Public Policy and Management, Tsinghua University; the Center for Forecasting Science, the Academy of Mathematics and System Science of the Chinese Academy of Sciences; the School of Economics and Management, University of Chinese Academy of Sciences; Paula and Gregory Chow Institute for Studies in Economics, Xiamen University; and the Basic Science Center for Econometric Modeling and Economic Policy Studies, Xiamen University–Chinese Academy of Sciences for their helpful suggestions and comments. The usual disclaimer applies.}}

\author{Yuzhi HAO\thanks{Department of Economics, The Hong Kong University of Science and Technology. Email: yuzhi.hao@connect.ust.hk.} \and Danyang XIE\thanks{Thrust of Innovation, Policy, and Entrepreneurship, the Society Hub, The Hong Kong University of Science and Technology (Guangzhou). Email: dxie@hkust-gz.edu.cn.}}

\date{February 23, 2025}

\begin{document}
\begin{CJK*}{UTF8}{gbsn} 
\newgeometry{top=2cm}
\thispagestyle{empty}  

\maketitle
\begin{minipage}{0.96\textwidth}
\begin{abstract}
This paper pioneers a novel approach to economic and public policy analysis by leveraging multiple Large Language Models (LLMs) as heterogeneous artificial economic agents. We first evaluate five LLMs' economic decision-making capabilities in solving two-period consumption allocation problems under two distinct scenarios: with explicit utility functions and based on intuitive reasoning. While previous research has often simulated heterogeneity by solely varying prompts, our approach harnesses the inherent variations in analytical capabilities across different LLMs to model agents with diverse cognitive traits. Building on these findings, we construct a Multi-LLM-Agent-Based (MLAB) framework by mapping these LLMs to specific educational groups and corresponding income brackets. Using interest-income taxation as a case study, we demonstrate how the MLAB framework can simulate policy impacts across heterogeneous agents, offering a promising new direction for economic and public policy analysis by leveraging LLMs' human-like reasoning capabilities and computational power.



\end{abstract}   
\end{minipage}
\restoregeometry
\newpage  
\section{Introduction}

The integration of Large Language Models into economic analysis offers new possibilities for policy research and computational economics. Traditional macroeconomic models, while mathematically rigorous, often rely on strong assumptions of rational expectations and representative agents. These models typically abstract complex economic processes through mathematical formulations such as production functions, which may not fully capture the rich dynamics of R\&D, human capital formation, and product evolution in real economies. Moreover, the inherent trade-off between model complexity and tractability has historically limited our ability to incorporate meaningful heterogeneity in agent behaviors and preferences with the exceptions such as \cite{aiyagari1994uninsured}, \cite{krusell1998income} and the TANK and HANK models of \cite{gali2004rule}, \cite{bilbiie2008limited}, and \cite{kaplan2018monetary}.

Agent-based models (ABMs) emerged as an alternative approach, allowing for the incorporation of heterogeneous agents with bounded rationality. Early ABMs, as documented by \cite{tesfatsion2006handbook} and examplified by \cite{brock1998heterogeneous}, relied on predetermined rules to govern agent behavior. While these models offered insights into emergent phenomena from individual interactions, they were constrained by the rigidity of their behavioral rules. Recent developments in artificial intelligence, particularly the advent of Large Language Models, offer new possibilities for enhancing the flexibility and sophistication of agent-based modeling.

A growing body of literature has begun to explore LLMs' potential in economic analysis. \cite{mei2024} and \cite{ma2024} demonstrate that LLMs exhibit human-like behavioral characteristics in economic decision-making scenarios. \cite{xie2024} shows that different AI chatbots exhibit distinct behavioral patterns in games, suggesting that these AI systems are not uniform in their decision-making patterns. \cite{chen2023} provides evidence of emergent economic rationality in LLMs' choices, while \cite{horton2023} evaluates their performance in classical behavioral economics games. In a pioneering application of LLMs to agent-based modeling, \cite{li2024} construct heterogeneous agents by varying the economic situations described in prompts to a single LLM, successfully simulating market dynamics and policy responses. However, their approach captures heterogeneity solely through varied prompts, while implicitly assuming uniform reasoning capabilities and behavioral tendencies across all agents.

Our paper advances this literature by introducing a two-dimensional approach to modeling heterogeneity. We capture differences not only in economic circumstances through varied prompts but also in reasoning capabilities and behavioral tendencies through the use of different LLMs. This innovation is motivated by the observation that economic decision-making heterogeneity in the real world stems from both objective circumstances (income, wealth, age) and subjective factors (cognitive abilities, cultural values, risk preferences). By employing multiple LLMs with distinct analytical capabilities and behavioral patterns, our framework better reflects the multifaceted nature of heterogeneity in real-world economic decision-making.

The potential of LLMs extends beyond individual decision-making to policy analysis. \cite{wu2024} explores how LLMs can help navigate trade-offs in policy research design, while \cite{pang2024} demonstrates their utility in enhancing causal analysis and research collaboration. These applications suggest that LLMs can serve not only as simulated economic agents but also as analytical tools for policy evaluation and institutional design. Our MLAB approach could therefore be particularly valuable for understanding how different population segments might respond to policy interventions.

Our paper makes several contributions to this emerging field. First, we provide a systematic evaluation of five leading LLMs' capabilities in solving dynamic economic optimization problems, using a canonical two-period consumption-savings framework. We assess these models under two distinct scenarios: one with explicit utility functions to test their computational abilities, and another without utility functions to examine their autonomous economic reasoning. Second, we introduce the Multi-LLM-Agent-Based (MLAB) framework, where different LLMs represent distinct socioeconomic groups based on their demonstrated capabilities. This novel approach leverages the varying characteristics of different LLMs to model population heterogeneity in economic decision-making.

We demonstrate the practical value of our MLAB framework through a case study of interest income taxation. By mapping different LLMs to specific educational backgrounds and corresponding income brackets in a calibrated population of 100 individuals, we simulate how various socioeconomic groups might respond to tax policy changes. Our results reveal nuanced patterns of behavioral responses that would be difficult to capture with traditional modeling approaches.

While previous research has simulated heterogeneity by solely varying prompts—akin to asking a highly intelligent agent to emulate the behavior of someone less educated—our approach harnesses the naturally diverse analytical capabilities of different LLMs to represent agents from distinct educational backgrounds. By mapping each LLM to a specific educational group, we enrich our model's representation of demographic and cognitive heterogeneity, thereby enhancing policy simulation exercises. Although the current assignment of LLMs to educational groups is not yet as scientifically rigorous or multidimensional as desired, our preliminary findings demonstrate the feasibility and robustness of this method. As the capabilities of LLMs continue to advance, we anticipate that a highly sophisticated LLM could be provided with detailed data on the characteristics of a target group and subsequently generate bespoke code to create a tailored LLM. This specialized LLM would then mimic the distinct “thinking and intuition” of the group, ultimately allowing for the creation of as many finely calibrated agents as necessary for comprehensive policy simulations. Such a methodological evolution underscores the transformative potential of the Multi-LLM-Agent-Based framework.

The remainder of the paper is organized as follows. Section 2 presents our theoretical framework and experimental design. Section 3 details our evaluation of LLMs' rational optimization capabilities. Section 4 examines their economic intuition and autonomous reasoning. Section 5 introduces the MLAB framework and presents our policy analysis results. Section 6 concludes with implications for future research and policy applications.

\section{Evaluating LLMs' Optimization Capabilities with Explicit Utility Functions}

We begin our analysis by evaluating LLMs' capabilities in solving a canonical two-period consumption-savings problem. This section presents our theoretical setting, parameter calibration, experimental design for assessing LLMs' rational optimization abilities when provided with an explicit utility function and results.

\subsection{Theoretical Setting}

Consider a representative middle-aged agent in urban China making consumption decisions over two distinct life periods of equal length (20 years each): a working period and a subsequent retirement period (e.g., ages 40-59 and 60-79). The agent's preferences are represented by a constant relative risk aversion (CRRA) utility function, widely adopted in macroeconomic analysis for its analytical tractability and empirical relevance. The optimization problem is specified as:

\begin{equation}
\max_{c_1,c_2} U = u(c_1) + \beta u(c_2)
\end{equation}

where

\begin{equation}
u(c) = \begin{cases}
\frac{c^{1-\sigma}}{1-\sigma}, & \text{if } \sigma \neq 1 \\
\ln(c), & \text{if } \sigma = 1
\end{cases}
\end{equation}

and the agent faces an intertemporal budget constraint:

\begin{equation}
c_1 + \frac{c_2}{(1+r)} = w_0 + y_1 + \frac{y_2}{(1+r)}
\end{equation}

where $c_1$ and $c_2$ represent consumption in periods 1 and 2 respectively, $\beta \in (0,1)$ is the subjective discount factor, $\sigma > 0$ is the coefficient of relative risk aversion, $r$ is the real interest rate, $w_0$ represents initial wealth, and $y_1$, $y_2$ denote income in each period.

The first-order conditions yield the Euler equation:

\begin{equation}
u'(c_1) = \beta (1+r) u'(c_2)
\end{equation}

Under CRRA utility, this becomes:

\begin{equation}
c_2 = c_1(\beta (1+r))^{\frac{1}{\sigma}}
\end{equation}

Combining with the budget constraint yields the analytical solutions:

\begin{equation}
c_1 = \frac{w_0 + y_1 + y_2/(1+r)}{1 + (\beta (1+r))^{\frac{1}{\sigma}}/(1+r)}
\end{equation}

\begin{equation}
c_2 = \frac{(w_0 + y_1 + y_2/(1+r))(\beta (1+r))^{\frac{1}{\sigma}}}{1 + (\beta (1+r))^{\frac{1}{\sigma}}/(1+r)}
\end{equation}

These solutions exhibit several key economic properties. First, consumption in both periods is proportional to lifetime wealth $(w_0 + y_1 + y_2/(1+r))$, reflecting the income effect. Second, the consumption ratio $c_2/c_1 = (\beta (1+r))^{1/\sigma}$ captures intertemporal substitution. Third, as $\sigma \to \infty$, consumption approaches perfect smoothing, while as $\sigma \to 0$, consumption becomes increasingly sensitive to interest rates.

\subsection{Parameter Calibration}

We calibrate our model parameters using data from the China Family Panel Studies (CFPS) 2018 wave, focusing on urban residents. For income parameters, we employ a two-step procedure that accounts for both age-specific income patterns and economic growth.

First, we divide the population into six age groups (20-29, 30-39, 40-49, 50-59, 60-69, and 70-79) and calculate the average income for each group ($\bar{y}_i$, where $i=1,\ldots,6$) and the overall population average income ($\bar{y}$). We then compute age-specific income ratios:

\begin{equation}
k_i = \frac{\bar{y}_i}{\bar{y}}, \quad i=1,\ldots,6
\end{equation}

To project future incomes, we assume a constant annual growth rate of 4\%. The average income $j$ decades from now, denoted as $\bar{y}^g_j$, is:

\begin{equation}
\bar{y}^g_j = \bar{y}(1.04)^{10j}, \quad j=0,\ldots,5
\end{equation}

For our representative agent who starts at age 20-29, their age-specific income in each decade is:

\begin{equation}
\bar{y}^{g,i} = \bar{y}^g_j k_i
\end{equation}

This formulation captures both life-cycle income patterns and economic growth. For instance, when our representative agent reaches ages 30-39, their income will be $\bar{y}^{g,2} = \bar{y}(1.04)^{10}k_2$.

To derive the two-period problem parameters, we calculate:

\begin{equation}
y_1 =  \bar{y}^{g,3}+\frac{\bar{y}^{g,4}}{(1+r)^{10}}
\end{equation}

\begin{equation}
y_2 =  \bar{y}^{g,5}+\frac{\bar{y}^{g,6}}{(1+r)^{10}}
\end{equation}

For initial wealth $(w_0)$, we solve a six-period optimization problem with identical preference parameters ($\beta$, $\sigma$) and interest rate $(r)$, starting with zero initial wealth at age 20. The optimal savings accumulated by the beginning of period 40-49 determines $w_0$.

This calibration yields the following parameter values:

Initial wealth $(w_0)$: 141,598.4 units

Working period income $(y_1)$: 958,189.8 units

Retirement period income $(y_2)$: 244,103.9 units

Discount factor $(\beta)$: $0.99^{20}$ (reflecting emerging market patience)

Risk aversion $(\sigma)$: 2 (standard macroeconomic calibration)

Annual interest rate $(r)$: 2\%

The relatively high discount factor reflects documented patience in emerging market economies, while the risk aversion parameter follows standard calibrations in macroeconomic literature. The interest rate combines information from borrowing rates (average real LPR 2019-2023, data from the People's Bank of China) and saving rates (real returns on wealth management products based on China Banking Wealth Management Market Annual Reports 2019-2023, adjusted using CPI data from the National Bureau of Statistics).

\subsection{Experimental Design}

Our experimental procedure evaluates five state-of-the-art LLMs: DeepSeek-V3, GPT-4o-20241120, Gemini-1.5-pro-002, Claude-3.5-sonnet-20241022, and Llama-3.1-405B. For each model, we conduct 16 independent trials to account for potential response variations. The experimental design follows a structured protocol to ensure consistency and reproducibility.

The prompt provided to each LLM is carefully designed to include three key components: role-playing setup, economic parameters, and output structure. The role-playing component establishes the decision-making context:

\begin{quote}
\itshape
Please get into the role-play mode. Imagine you are a middle-age working adult in an urban area of China, facing a consumption decision. You need to plan your consumption for the rest of your life, which can be divided into two periods with equal and large number of years: Working period followed by Retirement Period.
\end{quote}

The economic parameters section presents the utility function, budget requirement, and all relevant numerical values:

\begin{quote}
\itshape 
Your preferences over consumption in these two periods can be described by: $U = [c1^{(1-2)} + 0.818*c2^{(1-2)}]/(1-2)$ where: c1 is your consumption for the working period. c2 is your consumption for the retirement period.

Your current economic situation: Current savings: 141,598.4 units. Income for the working period: 958,189.8 units. Income for the retirement period: 244,103.9 units. Interest rate between periods: 48.6\%. You need to live within your means.
\end{quote}

The output structure requires a specific format for the response:

\begin{quote}
\itshape 
Given these circumstances, how would you choose your consumption for these two periods of life? Please explain your choices. Toward the end, say ``Final Answer: I will choose to ... because ..."
\end{quote}

The experimental protocol consists of the following steps for each LLM:
1. Initialize a new chat session to ensure a clean context.
2. Present the standardized prompt. 
3. Record the complete model response.
4. Clear the conversation history.
5. Repeat steps 1-4 for a total of 16 independent trials.
6. Move to the next LLM and repeat the entire process.

To maintain independence across trials, we clear the conversation history before each new prompt. This ensures that each response is based solely on the current prompt rather than being influenced by previous interactions. All responses are recorded in a standardized format for subsequent analysis.

We maintain the default temperature settings for each LLM to preserve their natural reasoning patterns:
\begin{quote}
Deepseek-V3: temperature = 1.0 in a 0-2 range\\
GPT-4o: temperature = 1.0 in a 0-1 range\\
Gemini-1.5-pro: temperature = 1.0 in a 0-2 range\\
Claude-3.5-sonnet: temperature = 1.0 in a 0-1 range\\
Llama-3.1-405B: temperature = 0.2 in a 0-1 range
\end{quote}
Interestingly, while Llama-3.1-405B has the lowest temperature setting, our subsequent analysis reveals that it exhibits the highest variation in decision-making patterns among all models.

\subsection{Performance Analysis and Model Comparison}

\begin{figure}[htbp]
    \centering
    \includegraphics[width=\textwidth]{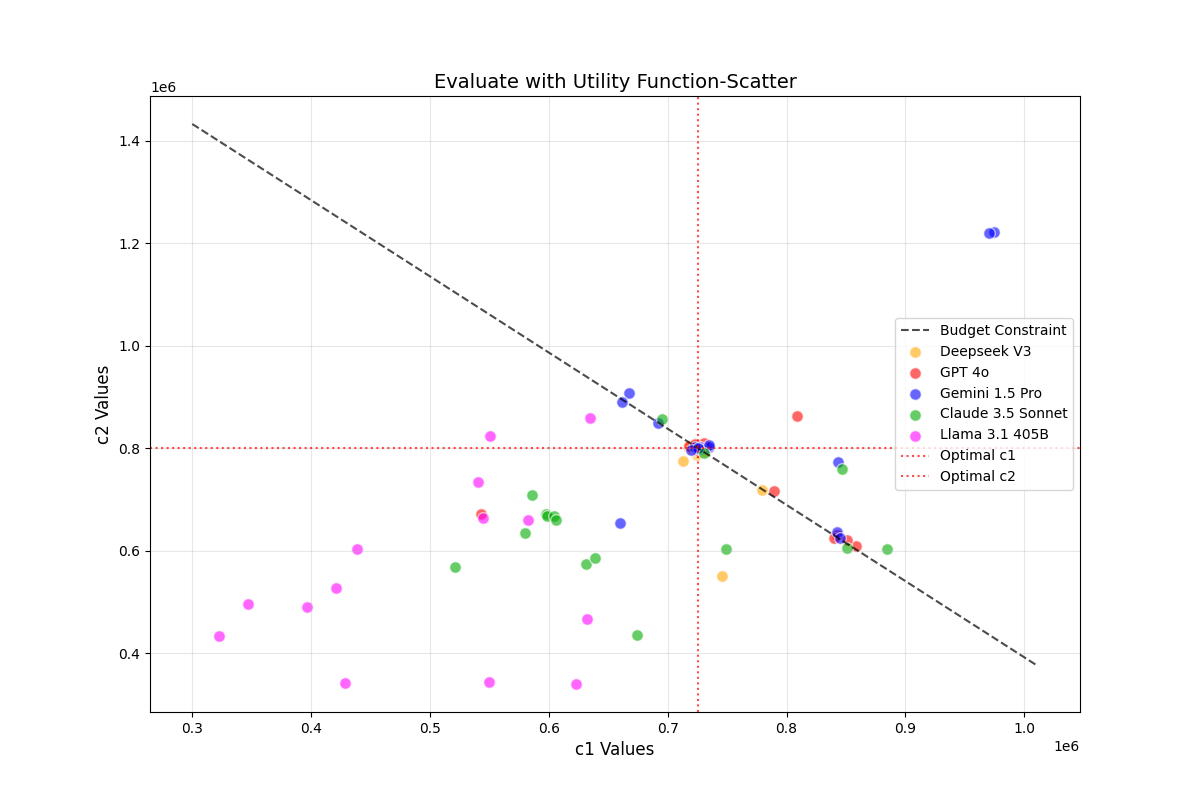}
    \caption{Consumption choices by LLMs. The diagonal line represents the budget constraint, and the dashed lines intersect at the theoretical optimal point. Each point represents one trial response.}
    \label{fig:scatter}
\end{figure}

Our analysis of the LLMs' responses reveals distinct patterns in their problem-solving capabilities. The results can be visualized in three complementary ways:

First, we plot each model's consumption choices (c1, c2) against the budget constraint line and the theoretical optimal point as shown in Figure~\ref{fig:scatter}. A point above the budget line indicates over-consumption (infeasible solution), while a point below represents under-consumption (inefficient resource utilization). The theoretical optimal point is marked by the intersection of two dashed lines.

DeepSeek-V3 demonstrates strong optimization capabilities, with most points clustering around the theoretical optimum. While some solutions fall slightly below the budget constraint line, indicating under-consumption, the majority of responses represent efficient resource allocation.

GPT-4o shows moderate performance, with points generally respecting the budget constraint but displaying greater dispersion around the optimal point. A few solutions appear above the budget line, indicating occasional failures to properly account for intertemporal resource constraints.

Gemini-1.5-pro exhibits more variable performance, with some solutions significantly exceeding the budget constraint. However, a notable cluster of points still appears near the optimal solution, suggesting partial understanding of the optimization problem.

Claude-3.5-sonnet frequently produces solutions below the budget constraint, indicating systematic under-consumption. The dispersion pattern suggests less consistent grasp of the optimization principles.

Llama-3.1-405B shows the weakest performance, with solutions widely dispersed and frequently far from both the budget constraint and optimal point. Many responses indicate severe under-consumption.

\begin{figure}[htbp]
    \centering
    \includegraphics[width=\textwidth]{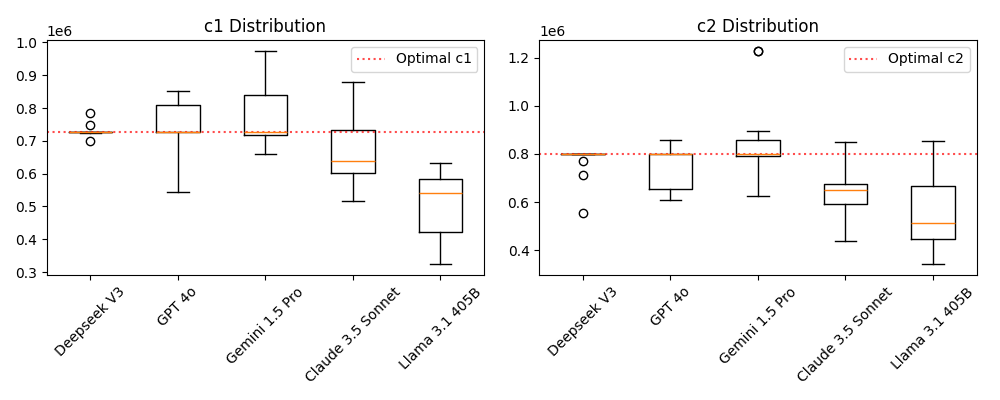}
    \caption{Distribution of consumption choices across periods.}
    \label{fig:distribution}
\end{figure}

The distribution of consumption choices across periods provides another perspective for analyzing model behavior. Figure~\ref{fig:distribution} presents the distribution of c1 and c2 for each model through box plots, where the boxes represent the interquartile range (25th to 75th percentiles), the middle line indicates the median, and the whiskers extend to the full range excluding outliers.

DeepSeek-V3 shows the most concentrated distributions for both c1 and c2, indicating consistent decision-making. In contrast, Llama-3.1-405B exhibits the widest dispersion, suggesting high variability in its recommendations even under identical conditions. GPT-4o and Gemini-1.5-pro display moderate variability, while Claude-3.5-sonnet shows a tendency toward lower consumption values with significant spread.

\begin{table}[h]
\caption{Performance comparison of different LLMs}
\begin{tabular}{lcccc}
\hline
Model & Accuracy & Mean Absolute & VarAPD & MAPD to \\
& (5\% tolerance) & Percentage & & Budget \\
& & Deviation & ($\times 10^4$) & Constraint \\
& (\%) & (\%) & & (\%) \\
\hline
Deepseek-V3 & 90.63 & 1.94 & 32.61 & 0.95 \\
GPT-4o & 56.25 & 7.33 & 81.75 & 1.96 \\
Gemini-1.5-pro & 50.00 & 10.95 & 214.76 & 6.68 \\
Claude-3.5-sonnet & 12.50 & 17.09 & 90.70 & 13.26 \\
Llama-3.1-405B & 3.57 & 31.35 & 266.06 & 30.66 \\
\hline
\end{tabular}
\label{tab:metrics}
\end{table}

The quantitative metrics, as shown in Table~\ref{tab:metrics}, confirm the visual patterns observed in the scatter plots. We evaluate model performance using four metrics: solution accuracy (percentage within 5\% of the theoretical optimum), Mean Absolute Percentage Deviation (MAPD) from optimal solutions, Variance of Absolute Percentage Deviation (VarAPD) for consistency assessment, and budget constraint adherence (measured by the percentage deviation between discounted consumption and total resources). DeepSeek-V3 demonstrates superior performance with 90.63\% accuracy and minimal deviations across all metrics. Performance decreases substantially across other models, from GPT-4o (56.25\% accuracy) to Llama-3.1-405B, which shows the largest deviations in both solution optimality and budget constraint satisfaction.

\section{Evaluating LLMs' Economic Intuition without Explicit Optimization Guidelines}
\subsection{Modified Prompt Design}

While the previous section evaluated LLMs' mathematical optimization capabilities, real-world economic decisions are rarely made through explicit utility maximization. Instead, individuals often rely on intuition, experience, and various economic and social considerations. This better reflects actual human decision-making processes, where choices are influenced by multiple factors beyond pure mathematical optimization.

To better simulate real-world decision-making, we modify our experimental design to assess LLMs' economic intuition without explicit optimization guidelines. The key modifications to the prompt focus on eliciting intuitive responses while maintaining the same economic scenario. Here is the complete prompt used in this section:

\begin{quote}
\itshape 
Please get into the role-play mode. Imagine you are a middle-age
 working adult in an urban area of China, facing a consumption decision.
 You need to plan your consumption for the rest of your life, which can be
 divided into two periods with equal and large number of years: Working
 period followed by Retirement Period. 
 
 Your current economic situation: Current savings: 141,598.4 units.
 Income for the working period: 958,189.8 units. Income for the retirement period: 244,103.9 units. Interest rate between periods: 48.6\%.
 You need to live within your means.
 Given these circumstances, how would you choose your consumption
 for these two periods of life based on your gut feeling? Please explain
 your choices.
 
 Toward the end, say ``Final Answer: I will choose to consume \_\_\_\_
 (specific number) of units during my working period and \_\_\_\_ (specific
 number) of units during my retirement period because\_\_\_\_\_”
\end{quote}

Compared to the previous section, we removed the utility function specification to avoid explicit mathematical optimization. Instead, we added the phrase ``based on your gut feeling" to encourage intuitive decision-making. We also included more specific formatting requirements for the final answer while maintaining identical economic parameters to ensure comparability between sections.

To maintain consistency and facilitate direct comparison with our previous findings, we followed the same experimental protocol as in Section 2. This includes conducting 16 independent trials for each model under identical economic parameters and testing conditions. We maintained the same testing environment and model versions throughout the experiments, employing consistent data collection and analysis methods across all trials.

\subsection{Behavioral Patterns and Decision Analysis}

\begin{figure}[htbp]
    \centering
    \includegraphics[width=\textwidth]{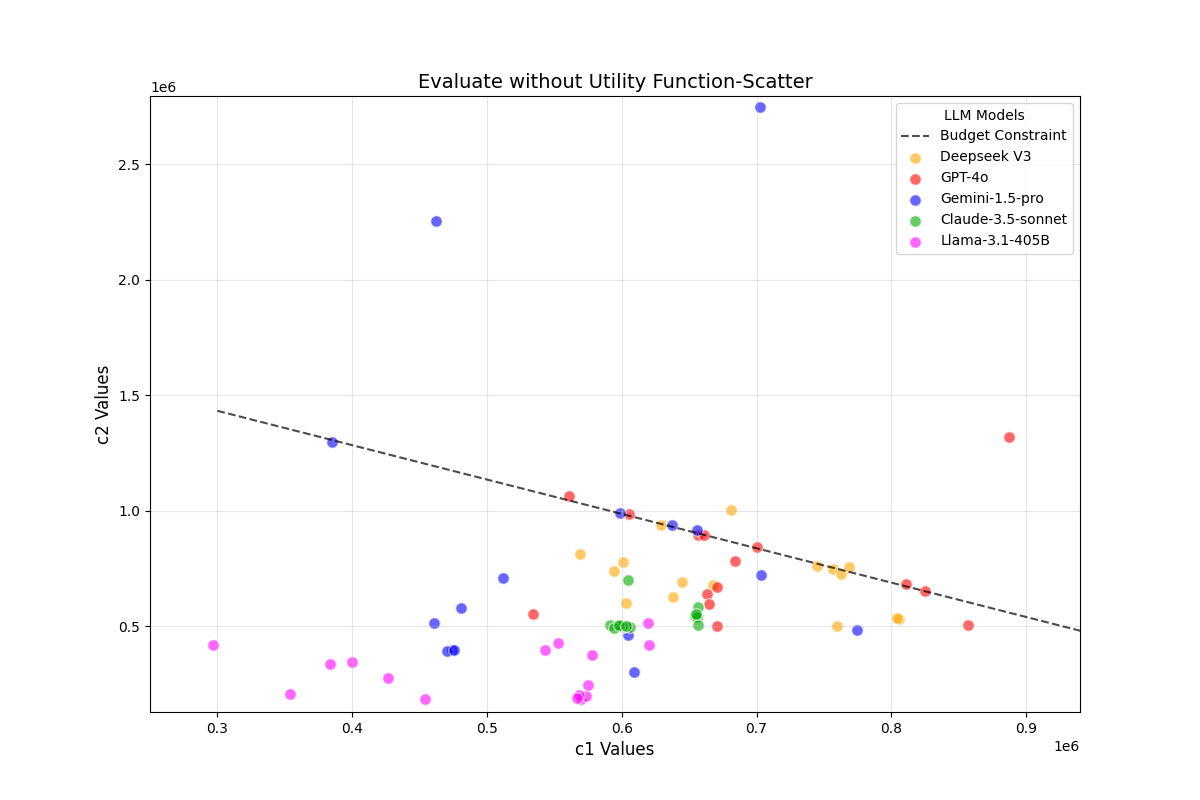}
    \caption{Consumption choices by LLMs without utility function specification. The diagonal line represents the budget constraint.}
    \label{fig:scatter_noutility}
\end{figure}

First, we examine the consumption choices without utility function guidance. Figure~\ref{fig:scatter_noutility} shows the scatter plots for each model.

Even without explicit optimization guidelines, most models maintain some degree of economic rationality, though with greater variation in their choices and a general tendency to underconsume relative to their budget constraints - a pattern more pronounced than in the utility function case. DeepSeek-V3 continues to show relatively concentrated choices, while Llama-3.1-405B exhibits the most dispersed pattern and the strongest tendency to leave resources unspent.

\begin{figure}[htbp]
    \centering
    \includegraphics[width=\textwidth]{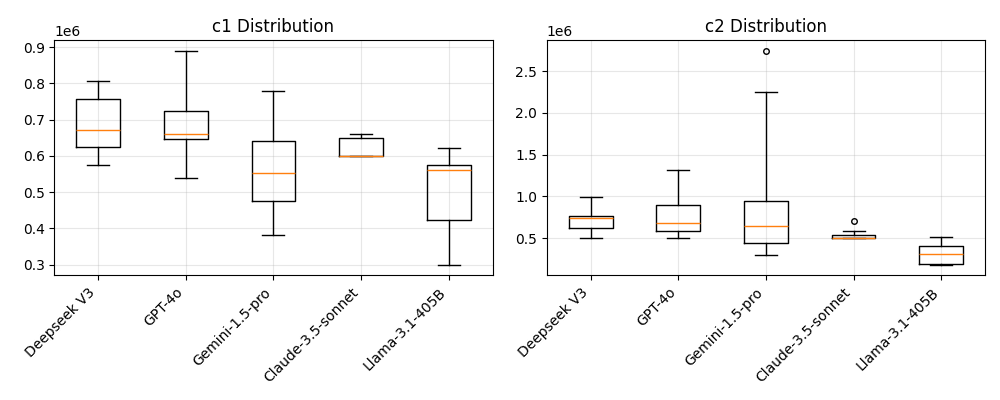}
    \caption{Distribution of consumption choices without utility function specification.}
    \label{fig:dist_noutility}
\end{figure}

The distribution analysis in Figure~\ref{fig:dist_noutility} reveals varying degrees of consistency across models. Claude-3.5-sonnet shows more concentrated responses compared to its performance with utility function, while other models display greater dispersion.

\subsection{Qualitative Analysis of Economic Decision-Making}

One significant advantage of using LLMs in economic analysis is their ability to not only make choices but also articulate their reasoning process. By analyzing the explanations provided across 16 trials for each model, we can better understand their decision-making patterns and behavioral tendencies. Below (Table~\ref{tab:motives}) we summarize the key economic motives mentioned in their responses:
\captionsetup{width=.95\textwidth}
\begin{longtable}{p{2cm}p{6cm}p{4cm}c}
\caption{Qualitative Analysis of Economic Decision-Making} \\
\hline
Model & Main Reasons & Economic Terms & Freq. \\
\hline
\endfirsthead
\caption{Economic Reasoning Analysis by Model (Continued)} \\
\hline
Model & Main Reasons & Economic Terms & Freq. \\
\hline
\endhead
\hline
\endfoot

\multirow{4}{=}{DeepSeek-V3} 
 & Balance between both periods & Consumption Smoothing & 16/16 \\
 & Save to take advantage of high interest rates & Intertemporal Substitution & 7/16 \\
 & Have a buffer during retirement & Precautionary Saving & 2/16 \\
 & Active lifestyle during working years & Higher Marginal Utility of Consumption & 1/16 \\
\hline
\multirow{7}{=}{GPT-4o} 
 & Balance between both periods & Consumption Smoothing & 16/16 \\
 & Save to take advantage of high interest rates & Intertemporal Substitution & 9/16 \\
 & Uncertainties like health issues in retirement & Precautionary Saving & 4/16 \\
 & Lower needs in retirement & Decreased Marginal Propensity to Consume & 3/16 \\
 & Rising living costs in retirement & Life-Cycle Spending Shifts & 2/16 \\
 & Peak consumption needs in working period & Life-Cycle Peak Consumption & 2/16 \\
 & Save against future income risks & Risk Aversion & 1/16 \\
\hline
\multirow{12}{=}{Gemini-1.5-pro}
 & Balance between both periods & Consumption Smoothing & 16/16 \\
 & Prices will go up & Inflation & 8/16 \\
 & Save to take advantage of high interest rates & Intertemporal Substitution & 5/16 \\
 & Chinese culture: emphasize saving & Risk Aversion & 4/16 \\
 & Work hard, deserve to consume in working period & Hedonic Consumption & 4/16 \\
 & Interest rates might change, could be a scam & Interest Rate Risk & 3/16 \\
 & Rising living costs in retirement & Life-Cycle Spending Shifts & 2/16 \\
 & Active lifestyle during working years & Higher Marginal Utility of Consumption & 2/16 \\
 & Uncertainties like health issues in retirement & Precautionary Saving & 2/16 \\
 & Left over for future generations & Bequest Motive & 2/16 \\
 & Lower needs in retirement & Decreased Marginal Propensity to Consume & 1/16 \\
 & Not being a burden on children & Parental Altruism & 1/16 \\
\hline
\multirow{10}{=}{Claude-3.5-sonnet}
 & Balance between both periods & Consumption Smoothing & 16/16 \\
 & Chinese culture: emphasize saving & Risk Aversion & 13/16 \\
 & Save to take advantage of high interest rates & Intertemporal Substitution & 12/16 \\
 & Rising living costs in retirement & Life-Cycle Spending Shifts & 10/16 \\
 & Peak consumption needs in working period & Life-Cycle Peak Consumption & 7/16 \\
 & Active lifestyle during working years & Higher Marginal Utility of Consumption & 5/16 \\
 & Uncertainties like health issues in retirement & Precautionary Saving & 4/16 \\
 & Lower needs in retirement & Decreased Marginal Propensity to Consume & 3/16 \\
 & Not being a burden on children & Parental Altruism & 3/16 \\
 & Help children in the future & Dynastic Utility Optimization & 1/16 \\
\hline
\multirow{6}{=}{Llama-3.1-405B}
 & Balance between both periods & Consumption Smoothing & 16/16 \\
 & Save to take advantage of high interest rates & Intertemporal Substitution & 9/16 \\
 & Lower needs in retirement & Decreased Marginal Propensity to Consume & 3/16 \\
 & Work hard, deserve to consume in working period & Hedonic Consumption & 2/16 \\
 & Rising living costs in retirement & Life-Cycle Spending Shifts & 1/16 \\
 & Life is short, and I don't know what the future holds & Present Bias & 1/16 %\hline
\label{tab:motives}
\end{longtable}

\subsection{Model Characteristics Analysis}

The analysis reveals both common patterns and distinct characteristics across models. All models demonstrate consistent emphasis on consumption smoothing, with all 16 responses across each model incorporating this fundamental economic principle. This uniformity suggests that LLMs exhibit human-like lifecycle consumption behavior patterns. Most models also frequently reference intertemporal substitution driven by the high interest rate, though with varying frequencies ranging from 5 to 12 responses, indicating widespread sensitivity to interest rate incentives. Additionally, precautionary saving motives appear across multiple models, although they manifest with different emphases and specific concerns, from health uncertainties to income volatility.

Beyond these commonalities, each model exhibits distinct characteristics in their decision-making approaches. DeepSeek-V3 demonstrates strong professional orientation through frequent use of economic terminology such as intertemporal budget constraint and consumption smoothing, maintaining the most rigorous economic reasoning framework throughout its responses. 

Gemini-1.5-pro displays unique critical thinking by questioning interest rate sustainability despite given conditions and, notably, introducing inflation concerns despite the prompt being framed in real terms. This tendency to suggest more conservative approaches based on additional considerations reflects a more comprehensive analytical approach to economic decision-making.

Claude-3.5-sonnet distinguishes itself through heavy emphasis on Chinese cultural values, frequently incorporating idioms like ``未雨绸缪  (prepare for rainy days)" and ``face" in its reasoning. This integration of cultural factors with economic decision-making provides a unique perspective on how social norms influence financial choices.

Llama-3.1-405B shows a distinctive pattern by consistently deviating from the Permanent Income Hypothesis, systematically under-consuming and reducing consumption due to lower retirement income. This behavior suggests less sophisticated economic reasoning but may reflect common behavioral biases observed in certain population segments.

These similarities in fundamental economic intuition, combined with distinct approaches to decision-making, make these models particularly suitable for representing different segments of the population in our MLAB framework. The variation in their reasoning processes helps capture the heterogeneity observed in real-world economic decision-making while maintaining basic economic rationality.

\section{A Multi-LLM-Agent-Based (MLAB) Framework}

\subsection{The Framework and Calibration}

The Multi-LLM-Agent-Based (MLAB) framework employs different LLMs to represent distinct population segments, offering a novel approach to modeling economic heterogeneity. Our framework innovatively captures two dimensions of heterogeneity: differences in economic circumstances through varied prompts, and variations in reasoning capabilities and behavioral tendencies through the use of different LLMs. This dual approach reflects the observation that economic decision-making heterogeneity in the real world stems from both objective circumstances (income, wealth, age) and subjective factors (cognitive abilities, cultural values, risk preferences).

Using the China Family Panel Studies (CFPS) 2018 data, we stratify the urban population aged 20-79 into five educational categories: 4-year college and above (including Master's and Doctoral degrees), 3-year college, senior high school/secondary school/technical school, junior high school, and primary school. For each educational category, we perform detailed parameter calibration following the methodology outlined in Section 2. The calibration process begins with extracting six 10-year period income patterns for each educational group. We then incorporate income growth projections and solve a six-period optimization problem, ultimately transforming the results into equivalent two-period economic parameters.

The mapping between LLMs and educational groups is designed to reflect both analytical capabilities and population shares. DeepSeek-V3 represents the highest educated group (11\% of the population), while GPT-4o corresponds to the upper-middle education segment (12\%). Gemini-1.5-pro represents the middle education bracket (24\%), Claude-3.5-sonnet models the lower-middle education segment (35\%), and Llama-3.1-405B represents the lowest education group (18\%). This mapping ensures that each LLM receives economic parameters calibrated to its corresponding educational group's income patterns, wealth levels, and lifecycle characteristics. The resulting framework thus captures both the economic heterogeneity arising from economic situation and the cognitive/behavioral heterogeneity inherent in different LLMs' decision-making processes.

\subsection{Case Study: Interest Income Taxation}

To demonstrate the MLAB framework's utility in policy analysis, we examine the impact of interest income taxation. The experiment introduces varying tax rates on interest earnings, ranging from 0\% to 100\%, to analyze how different segments respond to this policy intervention. This case study particularly highlights the value of our two-dimensional heterogeneity approach, as responses to tax policy may vary both due to different economic circumstances and different reasoning patterns.

The modified prompt for this analysis includes specific tax parameters while maintaining the core two-period consumption-savings framework:

\begin{quote}
\itshape
Your current economic situation:\\
Current savings: \{current\_savings\} units.\\
Income for the working period: \{working\_income\} units.\\
Income for the retirement period: \{retirement\_income\} units.\\
Interest rate between periods: 48.6\%.\\
Tax rate on interest earnings: \{tax\_rate\}\%, tax will be paid in the retirement period.
\end{quote}

Note that the economic parameters (\{current\_savings\}, \{working\_income\}, \{retirement\_income\}) vary across LLMs according to their mapped educational group's calibrated values.

\subsection{Heterogeneous and Adaptive Responses to Tax Policy Changes}

The analysis of saving rates reveals distinct behavioral patterns across educational-income groups. We consider two alternative definitions of the saving rate. The first measure, 1-c1/(w0+y1), includes both initial wealth and current income in the denominator, while the second measure, 1-c1/y1, more strictly considers only current period income. Here, c1 represents first-period consumption, w0 is initial wealth, and y1 is first-period income. As both measures yield qualitatively similar patterns, we present results for both specifications for completeness.

\begin{figure}[h]
\centering
\includegraphics[width=\textwidth]{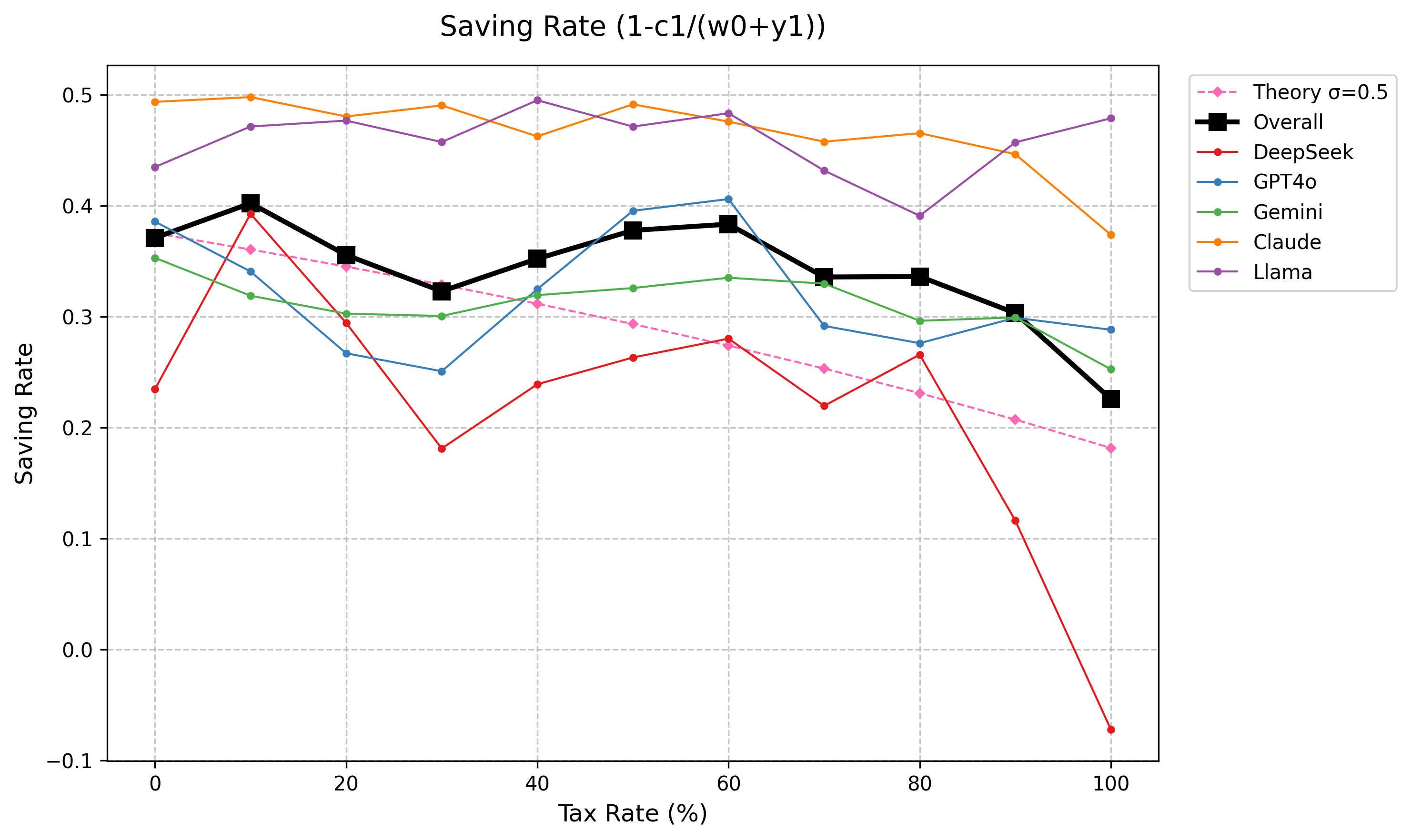}
\caption{Saving Rate Analysis (1-c1/(w0+y1))}
\label{fig:saving_rate1}
\end{figure}

\begin{figure}[h]
\centering
\includegraphics[width=\textwidth]{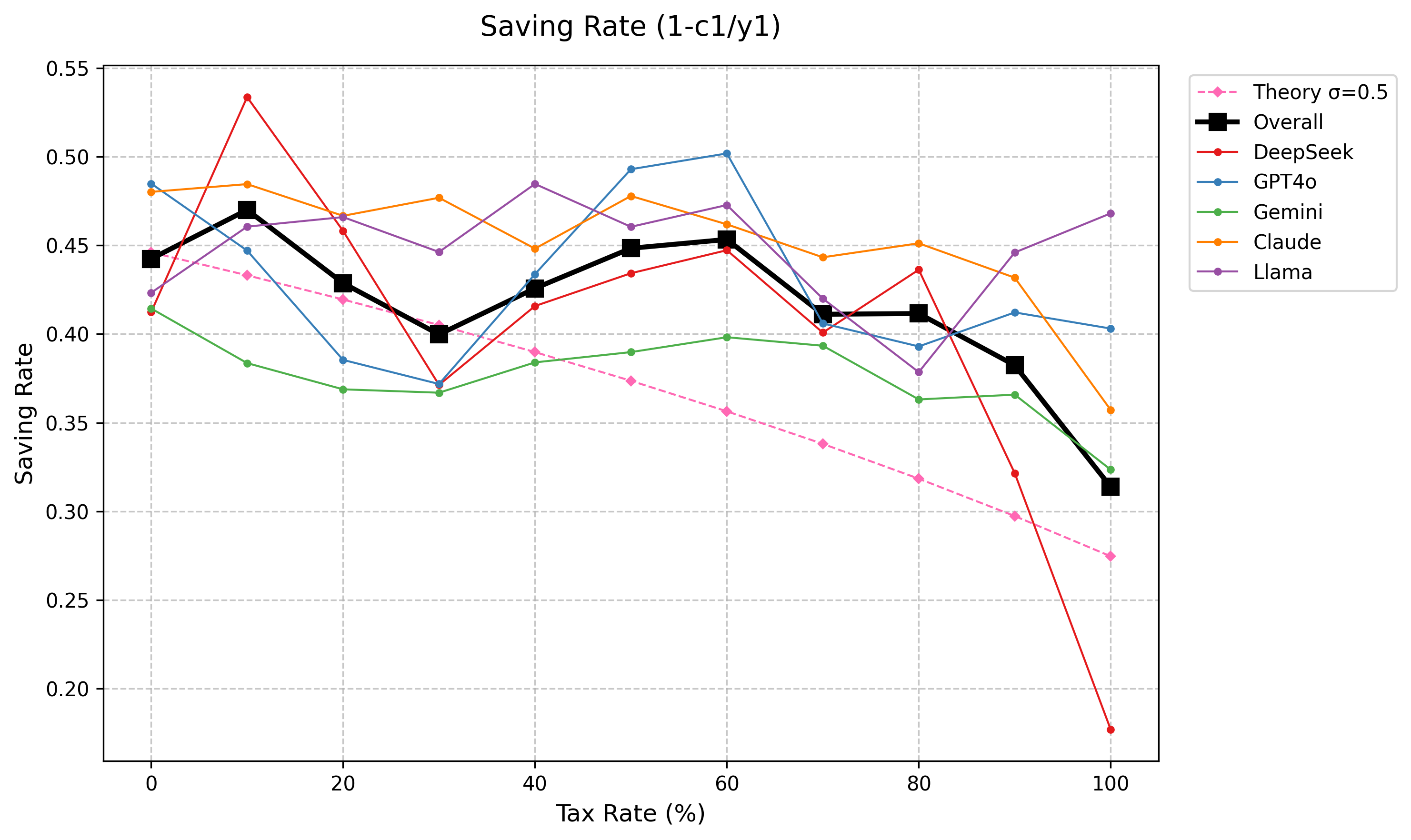}
\caption{Alternative Saving Rate Analysis (1-c1/y1)}
\label{fig:saving_rate2}
\end{figure}

For comparison purposes, we compute a reference saving rate path using the standard two-period optimization framework with CRRA utility. While $\sigma = 2$ is commonly used in macroeconomic calibrations, this parameterization generates a relatively flat saving rate profile (around 28\%) with a slight hump shape (not shown here), which fails to capture the magnitude of variation observed in our MLAB results, particularly the pronounced responses from models like DeepSeek-V3. We find that setting $\sigma = 0.5$ produces a monotonically declining saving rate profile that better approximates the observed pattern of tax sensitivity in our MLAB framework. This finding suggests that standard macroeconomic calibrations might understate the heterogeneity and magnitude of behavioral responses to tax policy changes.

The empirical responses from different LLMs exhibit varying degrees of volatility and sensitivity to taxation across both specifications, as shown in Figure~\ref{fig:saving_rate1} and Figure~\ref{fig:saving_rate2}. Notably, while the present paper employs a pragmatic mapping between LLMs and educational groups, the results affirm that even a relatively coarse alignment can capture meaningful behavioral distinctions across populations. The highest-educated group, represented by DeepSeek-V3, and the upper-middle education segment, represented by GPT-4o, show the most pronounced tax sensitivity. Their saving rates display significant variation across tax regimes, suggesting sophisticated tax planning and strong response to fiscal incentives.

The middle-education representation through Gemini-1.5-pro exhibits more stable saving patterns, indicating a balanced approach to consumption-savings decisions regardless of tax changes.  
The lower-education groups, proxied by Claude-3.5-sonnet and Llama-3.1-405B, demonstrate consistently conservative saving behavior, with less responsiveness to tax rate changes. 

The aggregate saving rate exhibits a general declining trend with increased taxation, albeit with greater volatility than predicted by standard models with homogeneous agents. This heterogeneity in responses, stemming from both economic circumstances and cognitive/behavioral differences, provides valuable insights for policy design, suggesting that tax policies may have differential impacts across educational and income groups. Moreover, the need to adjust the risk aversion parameter away from standard macroeconomic calibrations to match our MLAB results highlights the potential limitations of conventional representative agent models in capturing the full range of behavioral responses to policy changes.

\section{Conclusion and Future Research}

This paper introduces the Multi-LLM-Agent-Based (MLAB) framework, a novel pathway for enhancing policy simulations by leveraging the heterogeneous reasoning patterns inherent in different LLMs. Our research demonstrates that LLMs can effectively simulate human-like economic decision-making, exhibiting key behavioral patterns such as consumption smoothing, intertemporal substitution, and precautionary saving motives. The framework's innovation lies in its two-dimensional approach to heterogeneity, capturing both economic circumstances and cognitive capabilities through different LLMs.

Traditional economic models face limitations in addressing individual behavioral heterogeneity and policy simulation flexibility. The MLAB framework addresses these challenges by integrating multiple LLMs to simulate economic decisions across different income groups more flexibly. Our evaluation of LLMs in simulating individual consumption-saving behavior reveals their capacity to demonstrate human-like economic characteristics. The case study of interest income taxation policy demonstrates the model's ability to capture adaptive responses across different tax rates and income groups. Notably, while the present paper employs a pragmatic mapping between LLMs and educational groups, the results affirm that even a relatively coarse alignment can capture meaningful behavioral distinctions across populations, suggesting broader applications in policy analysis. 

Future research directions focus on several key areas for enhancement. First, the incorporation of more diverse LLMs could provide greater representation of population heterogeneity, potentially capturing additional demographic and socioeconomic segments. Second, developing dynamic agent interactions across income groups would enable more sophisticated modeling of economic behaviors and policy responses. This could include peer effects, social learning, and market interactions that influence individual decision-making.

The framework's methodology could be extended to analyze broader policy scenarios. For instance, in healthcare policy design, the MLAB framework could simulate how different population segments might respond to insurance market reforms or public health initiatives. In educational policy, the framework could evaluate the effectiveness of various subsidy schemes across different socioeconomic groups. Housing policy analysis could benefit from MLAB framework' ability to capture heterogeneous responses to mortgage market regulations and housing subsidies.

Looking ahead, as LLM technology evolves, it becomes conceivable to endow a highly capable LLM with specific demographic and behavioral data to automatically generate tailored LLM instances. These bespoke agents would be finely tuned to emulate the precise cognitive styles and decision-making processes of target groups, thus offering an unprecedented level of granularity in policy analysis. The MLAB methodology, therefore, holds substantial promise not only for its immediate applications but also as a foundation for future, more sophisticated adaptive agent-based models.

In conclusion, the MLAB framework represents a significant methodological innovation in economic modeling and policy analysis. By leveraging the capabilities of multiple LLMs to capture heterogeneous economic behaviors, it provides policymakers with a more nuanced tool for understanding policy impacts across different population segments. As both LLM technology and economic modeling techniques continue to advance, the MLAB framework stands to become an increasingly powerful instrument for policy design and evaluation, offering new perspectives on how different segments of society might respond to various policy interventions.
\newpage 
%\bibliography{references}

\end{CJK*}
\end{document}